\documentclass{article}

\usepackage{PRIMEarxiv}

\usepackage[utf8]{inputenc} 
\usepackage[T1]{fontenc}    
\usepackage{hyperref}       
\usepackage{url}            
\usepackage{booktabs}       
\usepackage{amsfonts}       
\usepackage{nicefrac}       
\usepackage{microtype}      
\usepackage{lipsum}
\usepackage{fancyhdr}       
\usepackage{graphicx}       
\graphicspath{{media/}}     

\usepackage{times}

\usepackage{soul}
\usepackage{url}
\usepackage[utf8]{inputenc}
\usepackage[small]{caption}
\usepackage{graphicx}
\usepackage{amsmath}
\usepackage{booktabs}

\usepackage{xcolor}

\usepackage{float}

\usepackage{graphicx}
\usepackage{textcomp}
\usepackage{xcolor}
\usepackage[ruled,vlined,linesnumbered,noend]{algorithm2e}
\usepackage{amsmath}
\usepackage{hyperref}       
\usepackage{url}            
\usepackage{booktabs}       
\usepackage{amsfonts}       
\usepackage{nicefrac}       
\usepackage{dblfloatfix}    
\usepackage{subcaption}
\usepackage{placeins}

\DeclareMathOperator{\Min}{min}
\DeclareMathOperator{\Max}{max}
\DeclareMathOperator{\Med}{med} 

\DeclareMathOperator{\abs}{abs}

\DeclareMathOperator{\Randint}{randint}

\DeclareMathOperator{\ReturnV}{return}

\DeclareMathOperator*{\argmin}{arg\,min}
\DeclareMathOperator*{\argmax}{arg\,max}

\title{AD-NEv: A Scalable Multi-Level Neuroevolution Framework for Multivariate Anomaly Detection
\thanks{\textit{}
\textbf{This work has been submitted to the IEEE TNNLS for possible publication. Copyright may be transferred without notice, after which this version may no longer be accessible.}} 
}

\author{
  Marcin Pietron \\
  Institute of Electronics \\
  AGH-UST \\
  Krakow\\
  \texttt{pietron@agh.edu.pl} \\
   \And
  Dominik Zurek \\
  Institute of Computer Science \\
  AGH-UST \\
  Krakow\\
  \texttt{dzurek@agh.edu.pl} \\
   \And
  Kamil Faber \\
  Institute of Computer Science \\
  AGH-UST \\
  Krakow\\
  \texttt{kfaber@agh.edu.pl} \\
   \And
  Roberto Corizzo \\
  Department of Computer Science \\
  American University \\
  Washington\\
  \texttt{rcorizzo@american.edu} \\
}

\begin{document}
\maketitle

\begin{abstract}
Anomaly detection tools and methods present a key capability in modern cyberphysical and failure prediction systems. Despite the fast-paced development in deep learning architectures for anomaly detection, model optimization for a given dataset 
is a cumbersome and time consuming process.
Neuroevolution could be an effective and efficient solution to this problem, as a fully automated search method for learning optimal neural networks, 
supporting both gradient and non-gradient fine tuning.
However, existing methods mostly focus on optimizing model architectures without taking into account feature subspaces and model weights.
In this work, we  propose Anomaly Detection Neuroevolution (AD-NEv) -- a scalable multi-level optimized neuroevolution framework for multivariate time series anomaly detection. 
The method represents a novel approach to synergically: \textit{i)} optimize feature subspaces for an ensemble model based on the bagging technique; \textit{ii)} optimize the model architecture of single anomaly detection models; \textit{iii)} perform non-gradient fine-tuning of network weights.
An extensive experimental evaluation on 
widely adopted multivariate anomaly detection benchmark datasets shows that the models extracted by AD-NEv outperform well-known deep learning architectures for anomaly detection. Moreover, results show that AD-NEv can perform the whole process efficiently, presenting high scalability when multiple GPUs are available. 
\end{abstract}

\keywords{Neuroevolution \and Anomaly Detection \and Autoencoders \and Neural Architecture Search}

\section{Introduction}
Modern cyberphysical and failure prediction systems involve sophisticated equipment that records multivariate time-series data from several up to thousands of features. Such systems need to be continuously monitored to prevent expensive failures. In anomaly detection, it is common to have abundant availability of normal data deriving from sensor monitoring, and scarcity of labeled anomalies. For this reason, most anomaly detection works focus on semi-supervised learning settings, where model training is conducted exclusively using normal data \cite{chalapathy2021}, \cite{pang2021}, as well as unsupervised learning settings, where training data is mostly normal but may contain a small number of unknown anomalies \cite{chalapathy2021}. Amongst recent works on semi-supervised and unsupervised multivariate time-series anomaly detection, deep learning-based methods achieve the best results on well known benchmarks \cite{malhotra2016lstmbased,zhou,zhang2019,su2019,USAD,tariq2019,tuor2017,zheng}. 

Within deep learning methods, a wide spectrum of autoencoder-based approaches were designed to deal with the anomaly detection problem. The most efficient are those based on convolutional, fully connected, and LSTM layers, or a combination of them in single model. Alternative methods are based on adversarial techniques \cite{USAD} as well as  variational autoencoders \cite{lstm-vae}
Other recent and very promising trends include autoencoders based on Graph Neural Networks \cite{graph-nn}, GAN-based architectures \cite{zhou}, supervised classification models \cite{nasa-lstm} and ensemble autoencoders \cite{garg}. 

In this context, one important problem is the identification of a suitable and optimized architecture for a given dataset, in a fully automated way.
Neuroevolution is a form of artificial intelligence that uses evolutionary approaches to find optimal neural networks. The most popular forms of neuroevolution algorithms include NEAT \cite{NEAT}, HyperNEAT \cite{stanley} and coDeepNEAT \cite{neuro_evolving}, which aim to optimize parameters, model architectures, or both.
However, despite its potential, neuroevolution approaches have been focused on the optimization of model architectures without taking into account the joint optimization of model architectures with feature subspaces and model weights. 

In this paper, we aim to fill this gap proposing AD-NEv, a multi-level neuroevolution approach that aims to identify robust and optimized autoencoder architectures for anomaly detection. Inspired by the framework formulated in \cite{faber2021}, which is loosely based on the coDeepNEAT algorithm, we consider two populations: models and subspaces. The former contains neural network architectures that evolve during the neuroevolution process. The latter consists of subspaces, which define subsets of input features. After the neuroevolution process, the framework sets up a bagging technique-based ensemble model from single optimised architectures. Moreover, our approach optimizes a single model for each subspace, overcoming the limitation of a single sub-optimal solution for all subspaces.
Subsequently, a fully scalable non-gradient fine-tuning process is performed to iteratively select the best solutions and generate model populations. Additionally, fine-tuning is performed on the evolved ensemble model. Our extensive experimental evaluation shows that our proposed multi-level neuroevolution approach yields deep ensembles of auto-encoder models that outperform state-of-the-art methods without requiring any predefined scoring function.

The main contributions of our work can be summarized in the following: 
\begin{itemize}
    \item A novel multi-level neuroevolution approach with a separate population of models for each subspace of features, which can be evolved independently, leading to a a better adaptation of specific models to each subspace.
    \item A novel selection process for models evolution based on an adapted distance measure for deep autoencoders that promotes model diversity.
    \item A fast non gradient-based fine-tuning approach for the evolved model architecture, leveraging adaptations of neural network weights in the evolution process, which improves results achieved by the previous levels.
    \item Automatic induction of a regularized ensemble model with a low number of sub-models, which further improves anomaly detection performance.
    \item An extensive evaluation with benchmark datasets that are widely adopted for multivariate anomaly detection.
\end{itemize}


\begin{figure*}[!b]
  \centering
  \includegraphics[width=0.90\linewidth,clip=true]{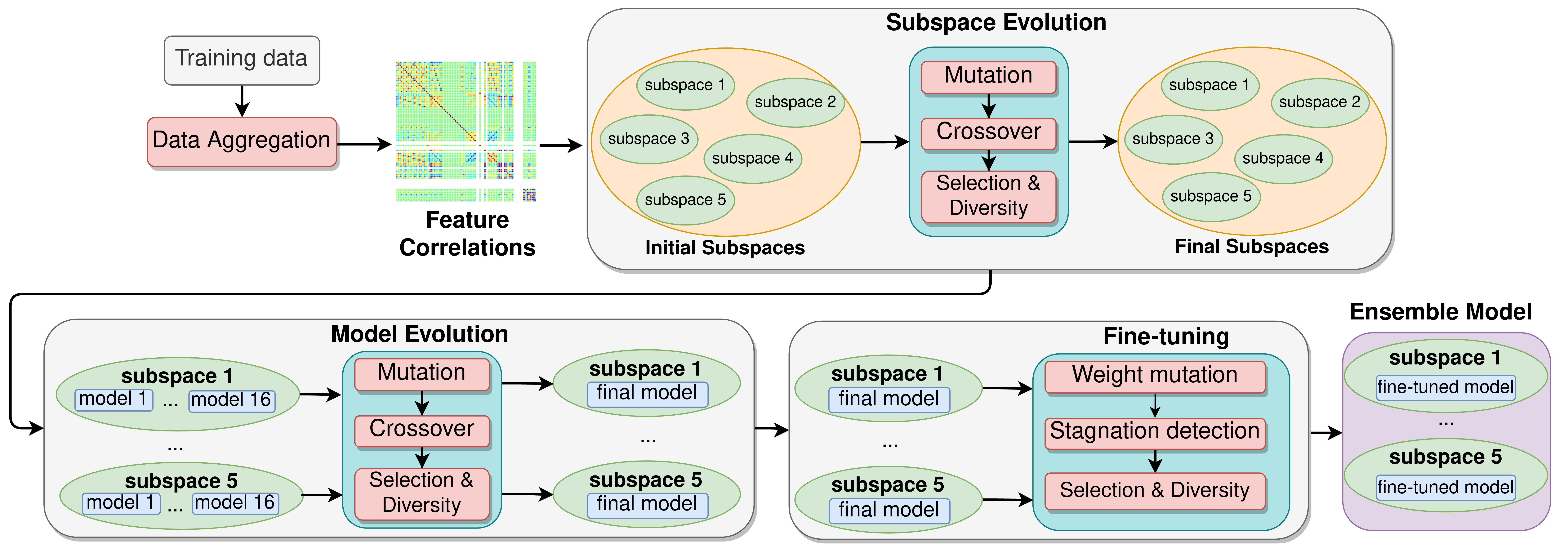}
    \caption{AD-NEv framework architecture. Time series data is used to train and evaluate a set of models during neuroevolution through different levels. The framework returns an optimized ensemble model that can be used for inference.}
    \label{fig:neral_schema}
\end{figure*}


The paper is organized as follows: Section \ref{sec:related_work}  describes related works. Section \ref{neuro-evolution} presents the proposed neuroevolution approach in detail. Section \ref{sec:experiments} describes and discusses our experiments. Section \ref{sec:results-discussion} summarizes the key results obtained in our study. Section \ref{sec:ablation_study} focuses on our ablation study. Finally, Section \ref{section:future_work} concludes the paper and outlines directions for future work.

\section{Related works}
\label{sec:related_work}
In this section, we analyze anomaly detection methods for multivariate time-series data, as well as neuroevolution methods that are most relevant for our research scope.
Recent surveys on general anomaly detection \cite{chandola2009}, deep learning based anomaly detection \cite{chalapathy2021}, \cite{pang2021} and unsupervised time-series anomaly detection \cite{blaz2020} present techniques relevant to unsupervised and semi-supervised multivariate time series anomaly detection. 
Autoencoder-based methods include fully-connected auto-encoder (FC AE),
USAD \cite{USAD}, and UAE \cite{garg}.
LSTM-based methods include NASA-LSTM \cite{nasa-lstm}, LSTM-AE \cite{garg} (which is based on \cite{malhotra2016lstmbased}), and LSTM-VAE \cite{Chen2019SequentialVF}. 
CNN-based methods include temporal convolutional AE (TCN AE) \cite{bai2018empirical}.
GAN-based methods include OCAN \cite{zheng}, and BeatGAN \cite{zhou}. Graph neural network-based approaches include \cite{graph-nn}.
Finally, hybrid approaches include MSCRED \cite{mscred}, DAGMM \cite{zong2018}, and OmniAnomaly \cite{su}.  

Regarding autoencoder-based approaches, the FC AE model introduced in \cite{garg} is similar to UAE, but it involves a single model over all the features, where the input sample is a vector resulting from the concatenation of time steps observed for all sensors. 
The USAD model is an auto-encoder with an additional discriminator model and loss extensions to boost the final scores.
In \cite{garg} authors present comparative studies on multivariate anomaly detection models. They describe Univariate fully-connected Auto-Encoder (UAE) as a model consisting of multiple auto-encoders, each connected by its input to a separate feature. Each encoder is a multi-layer perceptron with a number of nodes corresponding to the number of time steps (window size), and a reduced number of dimensions in the latent space by a factor of 2. The decoder is a mirror image of the encoder with $tanh$ activation. 
The resulting ensemble model outperforms many other deep learning architectures.

Focusing on LSTM-based approaches, NASA LSTM is a 2-layer LSTM model that uses predictability modeling, i.e., forecasting for anomaly detection \cite{nasa-lstm}. The LSTM-AE presented in \cite{garg} consists of single LSTM layer for each encoder and decoder. LSTM-VAE \cite{lstm--vae} models the data generating process from the latent space to the observed space using variational techniques. 

The CNN-based approach of temporal convolutional network AE (TCN AE) described in \cite{garg} is an architecture in which the encoder is built from a stack of temporal convolution (TCN) \cite{bai2018empirical} residual blocks. In the decoder, convolutions in the TCN residual blocks are replaced with transpose convolutions. The  study shows that the scoring function has a significant impact on the point-wise f1-score. 

GAN-based methods include OCAN \cite{zheng}, an end-to-end one-class classification method in which the generator is trained to produce examples that are complimentary to normal data patterns, which is used to train a discriminator for anomaly detection using the GAN framework. BeatGAN \cite{zhou} uses a Generative Adversarial Network framework where reconstructions produced by the generator are regularized by the discriminator instead of fixed reconstruction loss functions. 

The dense graph neural network approach presented in \cite{graph-nn} models the anomaly detection problem as a graph neural network, where each node represents a single feature, and edges allow to represent data exchanged between different nodes. The graph-based modeling capabilities represent a distinctive trait of this method, and were shown to yield a significant performance boost other state-of-the-art methods with very well known multivariate time series datasets.

Hybrid methods such as MSCRED \cite{mscred} learn to reconstruct signature matrices, i.e. matrices representing cross-correlation relationships between channels constructed by pairwise inner product of the channels. It is efficient in the case of long-term anomalies that are significantly out of the normal data distribution. Deep Autoencoding Gaussian Mixture Model \cite{zong2018} uses a deep autoencoder to generate a low-dimensional representation and reconstruction error for each input data point. The output of the autoencoder is further fed into a Gaussian Mixture Model (GMM). DAGMM jointly optimizes the parameters of the deep autoencoder and the mixture model in an end-to-end fashion. The joint optimization balances autoencoding reconstruction and density estimation of latent representation. The proposed regularization helps the autoencoder escape from less attractive local optima and further reduce reconstruction errors. OmniAnomaly \cite{su} is a stochastic recurrent neural network for multivariate time series. Its main idea is to capture the normal patterns of multivariate time series by learning their robust representations with key techniques such as stochastic variable connection and planar normalizing flow. Then, it reconstructs input data and uses the reconstruction probabilities to determine anomalies.

One common drawback of these methods is that they do not perform automatic model optimization, and therefore require a significant manual effort to identify and tune the right architecture for the right domain and dataset. 
This limitation may be solved by neuroevolution approaches, which have recently been used in many machine learning tasks for improving the accuracy of deep learning models and finding optimal network topologies \cite{neuroevol_overview}, \cite{danilo2016}, \cite{ming2018} and \cite{huang2022}. 

The authors in \cite{neuro_evolving} show that a two-level neuroevolution strategy scheme can outperform human-designed models in some specific tasks e.g. language modeling and image classification. This strategy is based on the co-deep NEAT algorithm with two optimisation levels: single sub-block optimisation and composition of sub-blocks to form a whole network. In \cite{huang2022}, a novel deep reinforcement learning-based framework is proposed for electrocardiogram time-series signal. The framework is optimized by neuroevolution algorithm. The authors in \cite{ming2018} present the framework of the self-organizing map-based neuroevolution solver by which the SOM-like network represents the abstract carpool service problem. The self organizing map network is trained by using neural learning and evolutionary mechanism. In \cite{danilo2016} the novel neuroevolution approach is described. The algorithm is incorporated with powerful representation which unifies most of the neural networks into one representation and with new diversity preserving method called spectrum diversity. The combination of spectrum diversity with a unified neuron representation enables the algorithm to either outperform or have similar performance with NeuroEvolution of Augmenting Topologies (NEAT) on five classes of problems tested. Ablation tests show the importance of new added features in the unified neuron representation.   
In \cite{dimanov2021} a novel neuroevolutionary method for optimization the architecture and hyperparameters of convolutional autoencoders. The hypervolume indicator in the context of neural architecture search is introduced. Results show that images can be compressed by a factor of more than 10, while still retaining enough information. In \cite{okada2017} it is shown that genetic algorithm could evolve autoencoders that can reproduce the data better as the manually created autoencoders with more hidden units. The experiments were performed on the MNIST dataset. The first approach of co-evolutionary neuroevolution-based multivariate anomaly detection system is presented in \cite{faber2021}. 
However, one substantial limitation is that optimization of subspaces and models occur separately, so that one model is optimized for all subspaces. This characteristic limits the capability of the neuroevolution process to optimize the model for each specific subspace, forcing the model to compromise in order to handle all subspaces simultaneously, and potentially resulting in a loss of anomaly detection accuracy. Moreover, the proposed method does not provide fine-tuning capabilities, which would provide the opportunity to further optimize the model and improve anomaly detection performance.
Fine-tuning is a quite popular technique for improving the accuracy of the pre-trained models. The most popular technique is gradient-based fine-tuning \cite{zhou2021}, \cite{ro2021}. The non-gradient approach is quite rare but can yield significant improvements as shown in \cite{nagel2020}. 
The presented method improves the accuracy of pretrained quantised models.

One common drawback of existing neuroevolution methods is that they typically optimize model architectures or model weights in isolation. Moreover, the majority of approaches that optimize model weights focus on shallow neural networks. Finally, to the best of our knowledge, there is no neuroevolution approach that jointly optimizes feature spaces, model architecture, and model weights. 

\section{Method}
\label{neuro-evolution}
In this section, we describe AD-NEv, a scalable multi-level neuroevolution framework that jointly addresses the aforementioned limitations of anomaly detection and neuroevolution methods.   
The starting point of the framework is data preparation -- which consists of downsampling training data used in the following evolution levels, and splitting it into overlapping windows, which reduces the computational cost of the following steps. The next step is finding the optimal partition of input features into subspaces, leading to an effective matching between features and models, as well as a reduced number of models in the final ensemble. After that, model evolution is performed for each subspace. As the next level, the best model for each subspace extracted from the previous level is fine-tuned using the non-gradient genetic optimization method. Subsequently, the ensemble model combines all fine-tuned models evaluating them using a voting mechanism. A visual representation of the framework architecture is shown in Figure \ref{fig:neral_schema}.  In the following, we describe all levels in detail in separate subsections. 




\subsection{Autoencoders}
AD-NEv leverages a deep neural network architecture consisting of autoencoders as base models for the neuroevolution process. Autoencoders are unsupervised learning models that learn a compressed representation of raw input data, which is then used to accurately reconstruct inputs. They consist of two parts: an encoder \textit{E} and a decoder \textit{D}. The encoder learns how to efficiently compress and encode the input data \textit{X} in a new representation with reduced dimensionality -- latent variables \textit{Z}. The decoder learns how to reconstruct the latent variables \textit{Z} back to their original shape. The model is trained to minimize the \textit{reconstruction loss}, which corresponds to minimizing the difference between the decoder's output and the original input data. It can be expressed as:
\begin{equation}
    \mathcal{L}(X, \hat{X}) = ||X - AE(X)||_2,
    \label{eq:autoencoder}
\end{equation}
where:
\begin{equation}
    AE(X) = D(Z),   Z = E(X).
\end{equation}

The autoencoder in our method can consist of various layers, e.g., fully connected layers and convolutional layers. Several variants of autoencoder architecture exists, including AAE (Adversarial AutoEncoder \cite{makhzani2015}), VAE (Variational AutoEncoder \cite{xu2018}) or DAE (Denoising AutoEncoder \cite{bengio2013}) but the general idea of reconstruction loss minimization is similar in all cases, eq. \ref{eq:autoencoder}. 
When dealing with time series data, the error is predicted at each time point (norm between input and multivariate reconstructed vector), as in  Equation \ref{eq:error}:

\begin{equation}
    Er_t = ||X_t - AE(x_t)||_2.
    \label{eq:error}
\end{equation}





\subsection{Data reduction} 
\label{subsec:data_preparation}
The algorithm starts by reducing the training data for the evolution process. Specifically, 
consecutive number of data points $\sigma$ are aggregated by averaging values for each feature, leading to a coarser time granularity. As a result, we are able to provide a fast evolution while retaining the most important information. The original training dataset $X$ contains $N$ samples, and each sample contains data from $M$ features, as shown in Equation \ref{eq:raw_dataset} and \ref{eq:input}: 

\begin{equation}
X = \{x_t, t \in {1,2,...,N}\}. 
\label{eq:raw_dataset}
\end{equation}

\begin{equation}
x_{t} = \{x_{t_i},  i \in \{1, 2, ..., M\}\}.  
\label{eq:input}
\end{equation}

The reduced dataset is annotated as $X_r$, whereas $\sigma$ is the reduction ratio parameter:
\begin{equation}
X_r = \{x_{r_t}, t \in {1,2,...,\frac{N}{\sigma}}\}. 
\label{eq:down}
\end{equation}

Single data points from $X_r$ are obtained according to Equation \ref{eq:red}, where $\Med$ is the median function:
\begin{equation}
x_{r_t} = \Med (\{ x_{t * \sigma}, x_{t * \sigma + 1}, ..., x_{(t+1) * \sigma} \}).
\label{eq:red}
\end{equation}

During this step, data points are grouped into overlapping windows used in the following steps. The rationale is based on the fact that, working with time-series data, an aggregated context can be more beneficial for the algorithm than a single data point. The window size $l_w$ can change during the evolution process, since it is one of the parameters subject to mutation.

\subsection{General evolution algorithm}
\label{subsec:general_evolution}

In our multi-level neuroevolution framework, we apply a genetic algorithm in three levels: subspaces, models, and non gradient fine-tuning.
Genetic operators such as crossover and mutation are different for each specific level, and the models' evolution and fine-tuning have a unique selection process. Nevertheless, the structure of the genetic algorithm is the same for all levels. For this reason, the general algorithm is introduced here, and details about each specific level are provided later on in the paper.


The genetic process starts with the generation of the initial population $P_0$. Each population contains $N_p$ solutions. After that, the single iteration is repeated $N_g$ times. Each iteration starts with creating offspring from the parents by the crossover operator. Next, the mutation operator mutates each solution from the offspring with the probability $p_m$. Subsequently, the fitness of each solution is calculated. In our approach, generation process works on the model architecture in the evolution of a single model, on subspaces in the subspace optimization, and on weight values during fine-tuning. The combination of these three levels in our framework allow us to explore and exploit a larger search space during model optimization.
A single genetic iteration finishes with the selection of solutions that form a new population. The result of the genetic process is the final population $P_{N_g}$.


\subsection{Subspaces evolution}
\label{subsec:subspaces_evolution}

The goal of the this level of the algorithm is to find an optimal partitioning of input features into subspaces. We define a subspace $G_i$ as a subset of the input features that is specific for each dataset, as shown in the following equation:
\begin{equation}
G_i(X) \subseteq \{X_0, X_1, X_2, ..., X_M\},
\label{eq:subspaces}
\end{equation}
where $X_i$ denotes data from the $i$-th input feature.

\setlength{\textfloatsep}{4pt}

\begin{algorithm}
\SetAlgoLined
\SetKwInOut{Input}{inputs}
\KwResult{$S'$--- solution created by crossover}
\KwIn{$S_1$---parent solution with $K$ subspaces}
\KwIn{$S_2$---parent solution with $K$ subspaces}
$S' = \{ \} $\;
\For{$i \in [0, 1, \dots, K]$} {
  $g_1 = S_{1_i}$ \tcp*{Subspace from $S_1$}     
  $g_2 = S_{2_i}$  \tcp*{Subspace from $S_2$}
  $g_{min} = \Min{(\Min{(g_1)}, \Min{(g_2)})}$\;
  $g_{max} = \Max{(\Max{(g_1)}, \Max{(g_2)})}$\;
  $\gamma = \Randint{(g_{min}, g_{max})}$ \tcp*{Split point} 
  $g' = \{ \} $\;
  \For{$\kappa$ $\in$ $g_1$} {
    \If{$\kappa < \gamma$} {
      $g' \gets g' \cup \kappa$\;
    }
  }
  \For{$\kappa$ $\in$ $g_2$} {
    \If{$\kappa > \gamma$} {
      $g' \gets g' \cup \kappa$\;
    }
  }
  $S' \gets S' \cup g'$ \;
}
$\ReturnV{S'}$
 \caption{Subspaces Crossover Algorithm}
 \label{alg:crossover}
\end{algorithm}

The partition $S$ of input features contains $K$ subspaces. There is no restriction on the frequency of the presence for a single input feature in subspaces, which means that it can be used in zero, one, or more subspaces.
Our method leverages the genetic algorithm to find the optimal partition of the input features into subspaces. A single gene provides information about a given feature being present in a subspace $G$. To perform subspace evolution we adopt the genetic operators defined in \cite{faber2021}: Crossover (Algorithm \ref{alg:crossover}), Moving Mutation (Algorithm \ref{alg:moving_mutation}), Vanishing Mutation (Algorithm \ref{alg:vanishing_mutation}) and Adding Mutation (Algorithm \ref{alg:new_features_mutation}).
\begin{algorithm}
\SetAlgoLined
\SetKwInOut{Input}{inputs}
\KwResult{$S'$--- solution created by mutation}
\KwIn{$S$---solution containing $K$ subspaces}
\KwIn{$P_m$---probability of a mutation}
$S' \gets \emptyset$ \;
\For{$i \in [0, 1, \dots, K]$} {
  sample $r$ from $\mathcal{N}(0,\,1)$\;
  \If{$r < P_m$}{
   sample $\kappa$ from $S_i$ \tcp*{Feature to move}
   $j = (i+1) \bmod K $ \tcp*{id of the next subspace in solution}
   $S'_j \gets S_j \cup \kappa $\;
   $S' \gets S' \cup S'_j$
   }
 }
 \Return{$S$}
 \caption{Subspaces Moving Mutation}
 \label{alg:moving_mutation}
\end{algorithm}

\begin{algorithm}

\SetAlgoLined
\SetKwInOut{Input}{inputs}
\KwResult{$S'$--- solution created by mutation}
\KwIn{$S$---solution containing $K$ subspaces}
$S' \gets \emptyset$ \;

\For{$S_i \in S$} {
  $S_i' = S_i$ \;
  \For{$\kappa \in S_i$} {
    $C_\kappa = \sum_{S_i \in S}
    \begin{cases}
            1,  & \text{if } \kappa \in S_i \\
            0,  & \text{otherwise} \\
        \end{cases}$ \;
  sample $r$ from $\mathcal{N}(0,\,1)$\;
    \If{$r > \frac{1}{C_\kappa}$}{
        $S_i' \gets S_i' \setminus \{ \kappa\}$ \;
    }
  }
  $S' \gets S' \cup S_i'$ \;
}
\Return{$S'$}
\caption{Subspaces Vanishing Mutation}
\label{alg:vanishing_mutation}
\end{algorithm} 

\begin{algorithm}
\SetAlgoLined
\SetKwInOut{Input}{inputs}
\KwResult{$S'$--- solution created by mutation}
\KwIn{$S$---solution containing $K$ subspaces}
\KwIn{$F$--- set of all features}
$S' \gets \{ S_0', S_1', \dots, S_K' \},$ where $S_i' = S_i $ \;

\For{$\kappa \in F$} {
  \If{$\kappa \notin S$} {
      \For{$S_i \in S$} {
      sample $r$ from $\mathcal{N}(0,\,1)$\;
        \If{$r > \frac{1}{K}$}{
            $S_i' \gets S_i' \cup \kappa $\;
        }
      }
  }
}
\Return{$S'$}
 \caption{Subspaces Adding Feature Mutation}
 \label{alg:new_features_mutation}
\end{algorithm}

To improve the convergence speed of the genetic algorithm, we form the initial population based on the correlation between features, instead of using a randomly generated population. Features are clustered performing agglomerative clustering with a degree of randomness to achieve a diverse population. This process has the effect of recursively merging pair of clusters leveraging the linkage distance \cite{ackermann2014analysis}.  In our approach, we leverage correlations between features as an intuitive and automatic way to estimate their similarity and drive the clustering process.


\subsection{Models evolution}
\label{subsec:models_evolution}

This level follows subspace evolution and its aim is to find optimal models that can later be parts of the ensemble model. Models for each subspace are evaluated independently, since a model that is optimal for one subspace may not be efficient in another subspace. Therefore, we create and evaluate a single population of models for each subspace. We denote $P_{G_i}$ as a population of models specific for subspace $G_i$:
\begin{equation}
    P_{G_i} = \{ F_{\Theta}^j(G_i), j = {0, ..., N_P} \}.
\end{equation}

Each model is represented by an encoder and a decoder, as in Equation \ref{eq:model_e} and \ref{eq:model_d}, where the layers in encoder and decoder appear in reversed order:

\begin{equation}
z = E(F_{\Theta^E}^{j}(G_i)) = {f_{\theta_N^E}^{j}(f_{\theta_{N-1}^E}^{j}...(f_{\theta_{0}^E}^{j}(G_i)))},
\label{eq:model_e}
\end{equation}

\begin{equation}
G_i^r = D(F_{\Theta^D}^{j}(z)) = {f_{\theta_0^D}^{j}...(f_{\theta_{N-1}^D}^{j}(f_{\theta_{N}^D}^{j}(z)))}
\label{eq:model_d}
\end{equation}

Performing $D(E(F_{\Theta}^{j}(G_i)))$ we obtain the reconstructed subspace -- $G_i^r$. 


Each population $P_{G_i}$ is evolved independently from the others in order to find the best solution for each subspace $G_i$ by means of a genetic algorithm. The genetic operators follow the specifications in \cite{faber2021} and include crossover and mutation of the following parameters: number of layers, number of input and output channels for each layer $i$ ($F[i]_{ic}$, $F[i]_{oc}$), window size ($l_w$) (Algorithm  \ref{alg:model_mutate} and \ref{alg:model_cross})\footnote{Without loss of generality, Algorithms \ref{alg:model_mutate} and \ref{alg:model_cross} operate on any layer of the neural network. For conciseness, we do not report processing of the encoder and the decoder parts of the model.}. If a mutation of the number of layers takes place, the $L_F$ parameter is modified by removing the last $\Lambda_l$ layers in the encoder ($f_{\theta_N^E}^{j}$, $f_{\theta_{N-1}^E}^{j}$, ...) (lines 10-12, Algorithm \ref{alg:model_mutate}) and the first $\Lambda_l$ layers in the decoder ($f_{\theta_N^D}^{j}$, $f_{\theta_{N-1}^D}^{j}$, ...) or adding new ones to the end of the encoder after the $f_{\theta_N^E}^{j}$ layer, and to the decoder before $f_{\theta_N^D}^{j}$ (lines 14-17, Algorithm \ref{alg:model_mutate}). The mutation of the output channels in specific layer is described in lines 3-7, Algorithm \ref{alg:model_mutate}, the window size mutation is presented between lines 18-20.
In the case of a crossover, the chosen $l$\textit{-th} layer of the encoder $f_{\theta_l^E}$ and of the decoder $f_{\theta_l^D}$ are exchanged between two child models $F'_1$ and $F'_2$ in a subgroup population (lines 4-8, Algorithm \ref{alg:model_cross}). The next option of crossover is to exchange the length between two parent autoencoders (lines 10-18, Algorithm \ref{alg:model_cross}).


As the loss function $\delta$ for model $F$ and subspace $G$ on dataset $X$, we use the mean squared error defined as:
\begin{equation}
    \delta(F, G(X)) = \frac{1}{|G(X)|} \sum_{x \in G(X)} (F(x) - x)^2.
\label{eq:reconstruction_loss}
\end{equation}

\begin{algorithm}
\KwIn{F - a model}
\KwIn{$w_{max}$ - maximum window size, $L_{max}$ - maximum number of conv layers}
$F' \gets copy(F)$\;
sample $m$ from  $\{0,1,2\}$ \tcp*{mutation types}
\If{$m = 0$} { \tcp{mutate the number of channels in a layer} 
    sample $l$ from $\{0, \dots, L_{F}-1 \}$\;
    $c'$ = $randint(F[l]_{ic}, F[l+1]_{ic})$\;
    $F'[l]_{oc}$ $\gets$ $c'$\;
    $F'[l+1]_{ic}$ $\gets$ $c'$\;
}

\If{$m = 1$} { \tcp{reduce the length of the model}
    sample $l$ from $\{0, \dots, L_{max} \}$\;
    \uIf{$l < L_F$} {
    \For{$k \in [l+1,\dots,L_{F}]$} {
        $F'[k] \gets \emptyset$ \;
    }
    }
    \uElse{\For{$k \in [L_F,\dots,l-1]$} {
        $c'$ = $randint(F[k]_{oc}, F[k]_{oc}+c_{max})$\;
        $F'[k+1]_{ic}$ $\gets$ $F[k]_{oc}$\;
        $F'[k+1]_{oc}$ $\gets$ $c'$\;
    }}
    
}
\If{$m = 2$} { \tcp{mutate window size}
    sample $w$ from $\{1,\dots,w_{max}\}$\;
    $F'[0]_{ic}$ $\gets$ $w$\;
}
\Return{$F'$}

\caption{Single Model Mutation}
\label{alg:model_mutate}
\end{algorithm}

\begin{algorithm}
\KwIn{$F_1$, $F_2$}
$F_1' \gets copy(F_1)$\;
$F_2' \gets copy(F_2)$\;

sample $m$ from $\{0,1\}$ \tcp*{type of crossover}
\If{$m = 0$ \tcp*{exchange the layers} }{ 
sample $l$ from $\{0, 1, \dots, \Min{(L_{F_1}, L_{F_2}}) \}$ \;
$F_1'[l] $ $\gets$ $F_2[l]$\;
$F_2'[l] $ $\gets$ $F_1[l]$\;
}
\If{$m = 1$ \tcp*{exchange the lengths of models}} { 
    \uIf{$L_{F_1} > L_{F_2}$} {
        $F_1'[L_{F_2}-1]_{oc}$ $\gets$ $F_2[L_{F_2}-1]_{oc}$\;
        \For{$k \in [L_{F_2},\dots,L_{F_1}]$} {
            $F_1'[k]$ $\gets$ $F_2[k]$\;
            $F_2'[k]$ $\gets \emptyset$ \;
        }
    }
    \uElse {
        $F_2'[L_{F_1}-1]_{oc}$ $\gets$ $F_1[L_{F_1}-1]_{oc}$\;
        \For{$k \in [L_{F_1},...,L_{F_2}]$} {
            $F_2'[k]$ $\gets$ $F_1[k]$\;
            $F_1'[k]$ $\gets \emptyset\ $ ;
        }
    }
}
\Return{$F_1', F_2'$}
\caption{Models Crossover}
\label{alg:model_cross}
\end{algorithm}




The genetic algorithm needs to calculate the fitness $\Delta$ for each single model $F$ and subspace $G$. To achieve this goal, the methods relies on a windowed training dataset $X$ which is split into consecutive parts according to the timestamp of data points: a training part $X_t$ (80\%) and validation part $X_v$ (20\%). During the evolution, the model is trained on the training part $X_t$. After each evolution iteration, we calculate the fitness as the weighted loss from the validation datasets. The value is negated because the goal of the genetic algorithm is to maximise the fitness, whereas we want to minimise loss values. The whole calculation is expressed in the following equation:
\begin{equation}
    \Delta(F, G) = -  \frac{|X_t| * \delta(F, G(X_t)) + |X_v| * \delta(F, G(X_v))}{|X_t| + |X_v|}.
\label{eq:fitness}
\end{equation}

Our method introduces a novel selection process in the evolution of the models. Its goal is to avoid convergence to a local optimum by keeping diversity in the models' population. To achieve this goal, we modify the selection process. Instead of choosing only the best models in each generation, we also keep a few of the most different models. We calculate the distance $d_F$ between models $F_i$ and $F_j$. The value is based on the models' hyperparameters, such as the number of layers $L$ and the number of channels $\gamma$ in the convolutional or fully connected layer. The distance calculation is performed as:
\begin{equation}
    d_F(F_i, F_j) = L_{F_i} - L_{F_j} + \sum_{l_a, l_b \in L_{F_i}, L_{F_j}} \frac{\abs(\gamma(l_a) - \gamma(l_b))}{\min(\gamma(l_a), \gamma(l_b))},
    \label{eq:distance_div}
\end{equation}
where $\abs$ denotes the absolute value.

\subsection{Non-gradient fine-tuning}
\label{subsec:fine_tuning}
Changing weights values of pretrained models with gradient-based methods can result in sub-optimal models that could be further optimized.
This phenomenon was noticed and presented in \cite{nagel2020}, which shows that performing non-gradient fine-tuning after gradient-based optimization can yield more accurate models.
Inspired by this work, we perform non-gradient fine-tuning on the best models extracted from the previous level. 
As the optimization step, we choose the evolutionary approach in which the genetic operators modify weights values to improve the performance of the models. The mutation operator modifies the chosen weights by the mutation power $\tau$ with mutation probability $p_m$ as in:
\begin{equation}
    \theta' =  \theta * (1 \pm p_m*\tau),
    \label{eq:fine_tuning_mutation}
\end{equation}
where $\theta$ is a single weight.

\begin{algorithm}
\SetAlgoLined
\SetKwInOut{Input}{inputs}
\KwResult{$R$ - A set of fine-tuned models}
\KwIn{$F$ - The best model from previous level}
\KwIn{$N_{P}$ - Size of fine-tuning population in every generation}
\KwIn{$N_{g}$ - Number of generations in the fine-tuning}
\KwIn{$p_m$ - Mutation probability}
\KwIn{$\tau$ - Mutation power}
$R \gets \emptyset$ \;
$g \gets 0$ \tcp*{Generation number}
$F_g \gets F$ \;
\While{$g < N_{g}$} {
    $P_g \gets \{ F_{\Theta'_1}, F_{\Theta'_2}, \dots, F_{\Theta'_{N_P}} \}$ where $\theta'$ is created by mutating weights from $F_g$ using eq. \ref{eq:fine_tuning_mutation} with $p_m$ and $\tau$\;
    
    $F_g \gets \displaystyle \argmin_{F_i \in P_g} FP(F_i)$ \;

    \If{$FP(F_g) = 0  \lor \ $\\$
    \ \ \ \forall_{k \in [g-5, g-4, \dots, g-1]} FP(F_k) = FP(F_g)$ \tcp*{is stagnated}}{
        $R \gets R \cup F_g$ \; 
        $F_g \gets \displaystyle \argmax_{F' \in P} \ \  d_F(F, F') $ \tcp*{$d_F(\cdot, \cdot)$ from eq. \ref{eq:distance_fine_tuning}}
     }
    $g \gets g + 1$ 
}
\Return{$R$}
 \caption{Fine-tuning}
 \label{alg:fine}
\end{algorithm}

During the fine-tuning process formalized in Algorithm \ref{alg:fine}, the algorithm randomly mutates a percentage of the weights\footnote{To change weights, the plus/minus sign is randomly chosen with a probability of 50\%.} (line 5, Algorithm \ref{alg:fine}). The fitness for each solution is calculated in the population as the number of False Positives (FPs) (line 6, Algorithm \ref{alg:fine}). During the fine-tuning process, a sample is marked as an anomaly (FP) if its reconstruction error is higher then the average value of all reconstruction errors, multiplied by a constant factor that determines the allowed deviation from the mean.
The best solutions are selected to be included in the new population (lines 6-11, Algorithm \ref{alg:fine}). In the following iteration, a complete new set of models is generated based on the best models selected from the previous iteration.

If a fitness value of zero FPs is achieved (line 7, Algorithm \ref{alg:fine}), or when a model does not further improve for a given number of iterations, i.e. stagnation condition (line 8, Algorithm \ref{alg:fine}), no further improvement is possible. We recall that this assumption holds since model training takes place in a semi-supervised setting, in which training data contains no anomalies. 
If zero FPs are achieved or the stagnation condition occurs, the best model is removed from the population and is used for the inference phase, during which it is evaluated on testing data.

In all cases, after removing the best model, another model is selected (line 11, Algorithm \ref{alg:fine}) and further improved in the following iterations. The new model is selected based on the highest distance from the best model, as in the following equation:

\begin{equation}
    d_\theta(F_i, F_j) =  \sum_{k=0}^{Layers(F_i)}\sqrt{|\theta_{F_{ik}} - \theta_{F_{jk}}|^2}.
    \label{eq:distance_fine_tuning}
\end{equation}

The distance represents the difference between the value of the weights of the two models.
The process of selecting most diverse model aims at achieving the exploration of a larger space of models, while avoiding getting stuck in a local minima.

The fine-tuning process does not involve the crossover operation, since its adoption would provide the same effect as a very high value of the $p_m$ parameter. In that case, model weights would likely drift to a wrong direction in a single iteration, leading to a drastic loss in model's accuracy (as further showcased in Section \ref{sec:convergence_fine-tuning}).


\subsection{Ensemble model}
\label{subsec:ensemble_model}

The outcome of the previous steps can be summarized as:
\begin{itemize}
    \item An optimised partition of input features into subspaces: $\{G_i, i \in \{0, ..., K\} \}$.
    \item An optimised and fine-tuned model for each subspace $G_i$: $\{ F_{G_i}, i \in \{0, ..., K\} \}$.
\end{itemize}
Our next goal is to build an ensemble model that is capable to classify input data as normal or as anomaly. To accomplish this goal, we add a threshold to each models to return a binary prediction: either 0 (normal) or 1 (anomaly). Therefore, for each subspace $G_i$, we create a model $F'_{G_i}$ that compares the output of $F_{G_i}$ to the threshold $\eta_i$, specifically defined for each model:
\begin{equation}
    F'_{G_i}(x) = 
    \begin{cases}
    0   & F_{G_i}(x) < \eta_i \\
    1   & F_{G_i}(x) \geq \eta_i.
    \end{cases}
    \label{eq:classification_model}
\end{equation}
As a result of this process, we compose base models into the ensemble model $F_e$ adopting a crisp voting strategy:
\begin{equation}
    F_e(x) = \begin{cases}
    0   & \sum_{i = 0}^{K} F'_{G_i}(x) = 0 \\
    1   & \sum_{i = 0}^{K} F'_{G_i}(x) > 0. \\
    \end{cases}
\end{equation}

\begin{equation}
F_{e}(x) \iff (F_{G_1}(x_{G_1}), F_{G_2}(x_{G_2}), \ ... \ , F_{G_K}(x_{G_K})). 
\label{eq:ensemble}
\end{equation}

This voting mechanism was empirically determined and selected since it gives better results than standard approaches such as majority voting.
As a result, the ensemble model $F_e$ can classify input data as normal or an anomaly based on the output of single models.





\section{Experiments}
\label{sec:experiments}
Our experiments focus on multivariate anomaly detection with time series data, which represents a challenging task that is crucially important in many real-life applications.

More specifically, our experiments aim to address the following research questions: 

\begin{itemize}
    \item \textbf{RQ1:} Does AD-NEv yield a higher anomaly detection performance than state-of-the-art anomaly detection methods for multivariate time series? 
    \item \textbf{RQ2:} Are all modeling levels in AD-NEv contributing to the achievement of the final anomaly detection performance of the method? 
   \item \textbf{RQ3:} Does AD-NEv model training present a reasonable time complexity in terms of execution time for all its levels (training, subspaces, models, fine-tuning)?
   \item \textbf{RQ4:} Does the AD-NEv framework present satisfactory speedup values, allowing to efficiently scale model optimization when using multiple GPGPUs?
    \item \textbf{RQ5:} Does AD-NEv present a stable and efficient model evolution and fine tuning convergence process considering different values for the population size, $\tau$, and mutation probability parameters?

\end{itemize}

To answer these questions, we perform a set of quantitative and qualitative experiments that assess the performance of AD-NEv from different perspectives. In the following subsections, we describe the evaluation protocol followed in our study, the datasets analyzed, the metrics used to assess the anomaly detection performance, as well as competitor methods considered and their configuration. Results are discussed in Section \ref{sec:results-discussion}.

\subsection{Datasets}
\label{sec:datasts}
We selected relevant time series datasets, including the Secure Water Treatment Dataset (SWAT) \cite{SWAT}, the Water Distribution dataset (WADI) \cite{WADI}, the Mars Science Laboratory Rover dataset (MSL), and the Soil Moisture Active Passive Satellite dataset (SMAP) \cite{nasa-lstm}. All datasets either involve data gathered in real environments (MSL and SMAP) or carefully prepared testbeds reflecting existing systems (SWAT and WADI). The datasets contain anomalies resulting from either device malfunctions or external malicious activity. It is noteworthy that the chosen datasets are popular benchmarks used in many recent studies.\footnote{Extended descriptions and statistics of the datasets are available in the supplementary material.}

For all datasets, a clear separation in training and testing sets is provided. Training data contains clean conditions of the process without anomalies, whereas testing data contains both normal data and anomalies. As a result, our models are exclusively trained on normal data, without anomalies. Both training and testing data are processed in windows, according to the value of window size (in channels in the Conv1D layer) chosen during model optimization. Statistics about datasets characteristics are shown in Table \ref{statistc_of_datasets}.
    
\begin{table}[H]
\centering
\caption{Statistics of the used datasets}
\begin{tabular}{|c|c|c|c|c|}
\hline
\textbf{Datasets} & \textbf{Features} & \textbf{Training size} & \textbf{Test size} & \textbf{Anomalies} \\ \hline
\textbf{SWAT}     & 51                  & 49,668            & 44,981           & 11.97\%              \\ \hline
\textbf{WADI-2017}     & 123                 & 1,048,571           & 172,801          & 5.99\%               \\ \hline
\textbf{WADI-2019}     & 123                 & 784,571           & 172,801          & 5.77\%               \\ \hline
\textbf{SMAP}       & 55        & 138,004   & 435,826    & 12.82\% \\ \hline
\textbf{MSL}        & 27        & 58,317    & 73,729     & 10.48\% \\ \hline

\end{tabular}
\label{statistc_of_datasets}
\end{table}
    

While the SWAT and WADI datasets contain one continuous data stream, SMAP and MSL are more complex and they are split into smaller subsets called entities. We concatenate data from these subsets for the evolution of subspaces and models. Following this idea, to better fit specific data characteristics of entities, we build subspaces separately for each entity, and then train and fine-tune  separate models. The final result is calculated as the concatenation of the results from each entity. During prediction time, we choose the model corresponding to the particular entity of the source that generated the data instance.
    
\subsection{Metrics}
 The F1-score is a standard metric providing information about test accuracy. It is well-suited for anomaly detection tasks, as it is more resistant to class imbalance than other metrics such as Accuracy (which are susceptible to yield high values when the normal class in testing data is large and the anomaly class is underrepresented). We adopt F1 as a point-wise metric, i.e. we evaluate every data point independently. 
While sequence-based evaluation also represents a feasible choice and it is largely adopted in the literature, it usually results in higher values for performance metrics, without providing precise information about how well the algorithm can decide whether a single data point is an anomaly.


\section{Results and discussion}
\label{sec:results-discussion}

\subsection{Anomaly detection performance}
\begin{table}[]
\small
\centering
\caption{Point-wise $F_1$ score. * -- models were evaluated using Gauss-D scoring function; N/A -- results are not available. The best result for every dataset is in \textbf{bold}.}
\begin{tabular}{|l|c|c|c|c|c|}
\hline
\textbf{}   & \textbf{SWAT} & \textbf{WADI} & \textbf{MSL} & \textbf{SMAP}  & \textbf{Mean}\\ \hline 
\textbf{MAD-GAN}  & 0.77    &   0.37    &  N/A  & N/A & N/A \\
\textbf{GNN}      &  0.81   & 0.57  &  N/A &  N/A    & N/A \\      
\textbf{USAD}     & 0.79   & 0.43   &  N/A &  N/A   & N/A \\
\textbf{CNN 1D}   & 0.78   & 0.27  & 0.44 & 0.52 & 0.50 \\
\textbf{NASA LSTM*}  & 0.13  & 0.20 & 0.55 & 0.59 & 0.37 \\
\textbf{UAE*}  & 0.58 &  0.47  & 0.54 & 0.58    & 0.54          \\
\textbf{LSTM-AE*} &  0.45  & 0.33 & 0.54  & 0.53 & 0.46  \\      
\textbf{LSTM-VAE*} & 0.42 & 0.50 & 0.49 & 0.49 & 0.48 \\
\textbf{TCN AE*}     &  0.43 &  0.43  & 0.55 & 0.55 & 0.49 \\

\textbf{BeatGAN}     &  0.48 &  0.46  & 0.53 & 0.57 & 0.51
\\
\textbf{OCAN}     &  0.15 &  0.0  & 0.30 & 0.28 & 0.18
\\
\textbf{DAGMM}     &  0.0 &  0.13  & 0.14 & 0.17 & 0.11
\\
\textbf{OmniAnomaly}     &  0.15 &  0.24  & 0.41 & 0.38 & 0.29

\\ \hline
\textbf{AD-NEv}     &   \textbf{0.82} & \textbf{0.62}    & \textbf{0.57} & \textbf{0.77}  & \textbf{0.70}\\
\hline
\end{tabular}
\label{result_without_subspaces}
\end{table}

Table \ref{result_without_subspaces} presents the results for AD-NEv in comparison to the results of well-known anomaly detection methods. Some models are evaluated using an optimised Gauss-D scoring function \cite{garg}, which features a data transformation step on previously observed data points, aiming at increasing $F_1$ values. However, in our experiments, this scoring function did not provide significant improvements to model's accuracy and was therefore discarded in the final experiments.
For the largest dataset (WADI), AD-NEv outperforms the second-best method (GNN) by 0.05, whereas the $F_1$  of the third-best model (LSTM-VAE) is significantly lower, with a margin of 0.12. Most of the other methods obtain even worse results. The differences are less impressive with the SWAT dataset, where AD-NEv outperforms GNN and USAD by a value of 0.01 and 0.02, respectively. The smaller difference is probably related to the fact that SWAT is smaller and less complex than WADI, which results in methods being more aligned in terms of predictive performance. In the case of MSL, all the best models provide results that are very close to each other. However, AD-NEv improves the results of NASA LSTM and TCN AE by 0.02. 
However, AD-NEv provides a radical improvement with the SMAP dataset, where it outperforms the second-best model (NASA-LSTM) by 0.18.

Overall, it is noteworthy that the models extracted by AD-NEv achieve the best results across all datasets when compared with all the other methods.
For some datasets, the difference is very significant, while for others it is relatively smaller. However, the other methods considered in our experiments usually achieve high results with only few of the analyzed datasets. For example, NASA LSTM is the second-best method for MSL and SMAP datasets, but it does not handle SWAT and WADI very well (in which the $F_1$  is 0.13 and 0.20, respectively). On the other hand, AD-NEv showcases the capability to adjust the evolving model to specific dataset requirements thanks to the neuroevolution approach. This result can be clearly observed looking at the mean value of $F_1$  across all datasets in Table \ref{result_without_subspaces} (\textbf{RQ1}).


\begin{table}[]
\small
\centering
\caption{Ablation study showing point-wise $F_1$ score using a baseline model with basic subspace and model evolution and without fine tuning (A), using AD-NEv  with enhanced subspace and model evolution (B), and with the full version of AD-NEv that includes fine-tuning (C). The best result for each dataset is reported in \textbf{bold}.}
\begin{tabular}{|l|c|c|c|c|}
\hline
\textbf{Method} & \textbf{SWAT}     & \textbf{WADI}  & \textbf{MSL} & \textbf{SMAP}       \\\hline
\textbf{A: Baseline model \cite{faber2021}}   &  0.79 & 0.54 & 0.45 & 0.52 \\
\hline
\textbf{B: AD-NEv: Evolution}  & 0.81 & 0.59 & 0.50 & 0.67\\
\hline
\textbf{C: AD-NEv: Evolution +}   & \textbf{0.82} & \textbf{0.62} & \textbf{0.57}& \textbf{0.77}\\
\textbf{Fine-tuning}   &  &  & & \\\hline
\end{tabular}
\label{tab:result_with_subspace_and_evolution}
\end{table}

Table \ref{tab:result_with_subspace_and_evolution} presents results for an ablation study consisting of three models: \textit{i)} A baseline model \cite{faber2021}; \textit{ii)} After our subspaces and models evolution; \textit{iii)} After the fine-tuning level. This table reveals the improvements introduced by our newly introduced approaches for subspaces optimization, model evolution, and fine-tuning. Comparing the results from \cite{faber2021} with the results following our improvements in subspaces and model evolution, we see improvements in terms of $F_1$ on SWAT by 0.01, WADI by 0.05, MSL by 0.05, and SMAP by 0.15. These results show that the introduced novelties, such as the independent population for each subspace, improve the results with all datasets (\textbf{RQ2}). Fine-tuning also positively impacts all datasets: SWAT by 0.01, WADI by 0.03, MSL by 0.07, and SMAP by 0.10. The gains are lower for SWAT and WADI, as the results for these datasets are already higher. 
Those results confirm the positive contributions introduced by the new components proposed in this paper, and highlight that non-gradient fine-tuning can be an efficient method to boost the pretrained models' performance in anomaly detection tasks.




\subsection{Time Complexity and Scalability}
    \label{sec:execution_and_time}

    The framework was implemented\footnote{The implementation will be provided upon publication.} with Python and PyTorch. All presented calculations were executed leveraging Nvidia Tesla V100-SXM2-32GB GPUs\footnote{https://www.nvidia.com/en-us/data-center/v100/}. Since AD-NEv supports parallel model optimization, experiments were performed using a pool of 8 GPGPUs (each subspace/model was processed on a separate GPGPU). 
    
    \begin{figure}[h]
    \small
      \centering
      \includegraphics[scale=0.45]{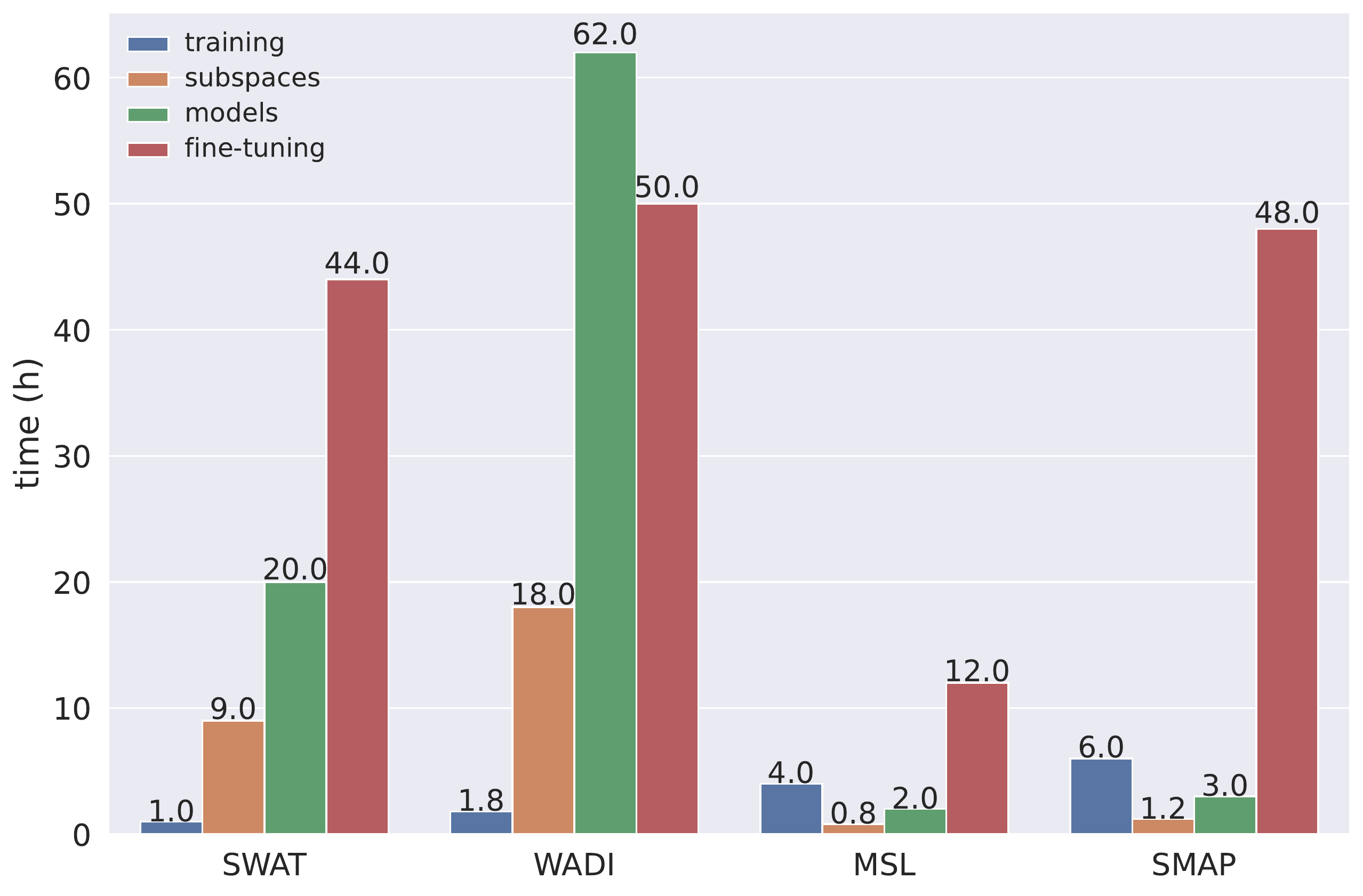}
        \caption{Execution times in hours using 8 GPGPUs with all datasets (SWAT, WADI, MSL, SMAP).}
        \label{fig:times}
    \end{figure}

    \begin{table}
\centering
\caption{Hyperparameters of the genetic process during each level}
\label{tab:genetic_params_all}
\begin{tabular}{|l|c|c|c|} 
\hline
\textbf{Parameter}    & \textbf{Subspaces}    & \textbf{Models}              & \textbf{Fine-tuning}                    \\ 
\hline
Population size       & 16                    & 24                           & 24                                      \\ 
\hline
Mutation probability  & 0.1                   & 0.5                          & 0.02                                    \\ 
\hline
Crossover probability & 0.1 & \multicolumn{1}{l|}{~~ 0.5} & \multicolumn{1}{l|}{not applicable}  \\ 
\hline
Number of generations & 10                    & 16                           & 64                                      \\
\hline
\end{tabular}
\end{table}

   \begin{table}[]
    \small
    \centering
    \caption{Other hyperparameters of the neuroevolution process}
    \label{tab:alg_params}
    \begin{tabular}{|l|c|}
    \hline
    {\textbf{Parameter}}   &  {\textbf{Value}} \\ \hline
    Reduction ratio  $\sigma$ (WADI, SWAT)    & 5 \\ \hline
    Reduction ratio  $\sigma$ (MSL, SMAP)    & 1 \\ \hline

    Mutation power $\tau$    & $\frac{1}{256}$ \\ \hline 
    Number of layers range &  [3-6] \\
    \hline
    Number of channels range &  [16-6144] \\
    \hline
    Window size range &  [1-12] \\
    \hline
    Learning rate range &  [0.000001-0.1] \\
    \hline
    \end{tabular}
    \end{table} 
    


    
    
    
 We provide the values of the parameters of the neuroevolution process for each level in Table \ref{tab:genetic_params_all} and \ref{tab:alg_params}. We also present the execution times in Figure \ref{fig:times}. It is important to note that SMAP and MSL datasets are processed in full, without downsampling, since most feature are binary, and this operation would results in information loss. Considering all datasets, the evolution of subspaces level took a minimum of 0.8 and a maximum of 18 hours. The best results in terms of accuracy were achieved with a value of the maximum number of possible subspaces (parameter $K$ of the evolution process) set to 5. The rationale for this choice was to find the highest number yielding a reasonable execution time. We experimentally observed that a higher value for $K$ exceeds the time limits of evolution process ($>$140h). Considering all datasets, the model evolution process took a minimum of 2 and a maximum of 62 hours. Fine-tuning was on average more time consuming, taking between 12 and 50 hours. It is noteworthy that model training and fine-tuning were performed separately for each entity in the SMAP and MSL datasets. Therefore, the whole neuroevolution process, including the final training and evaluation, lasted the following execution times: SWAT -- 74h; WADI -- 131.8h; MSL -- 18.8h; SMAP -- 58.2h.    
Overall, all the executions for each single dataset required less than 132 hours, which we consider to be a reasonable time frame considering the size of the datasets and the satisfactory accuracy achieved by the resulting models (\textbf{RQ3}).  
     
\begin{figure}[H]
\small
\centering
\includegraphics[scale=0.6]{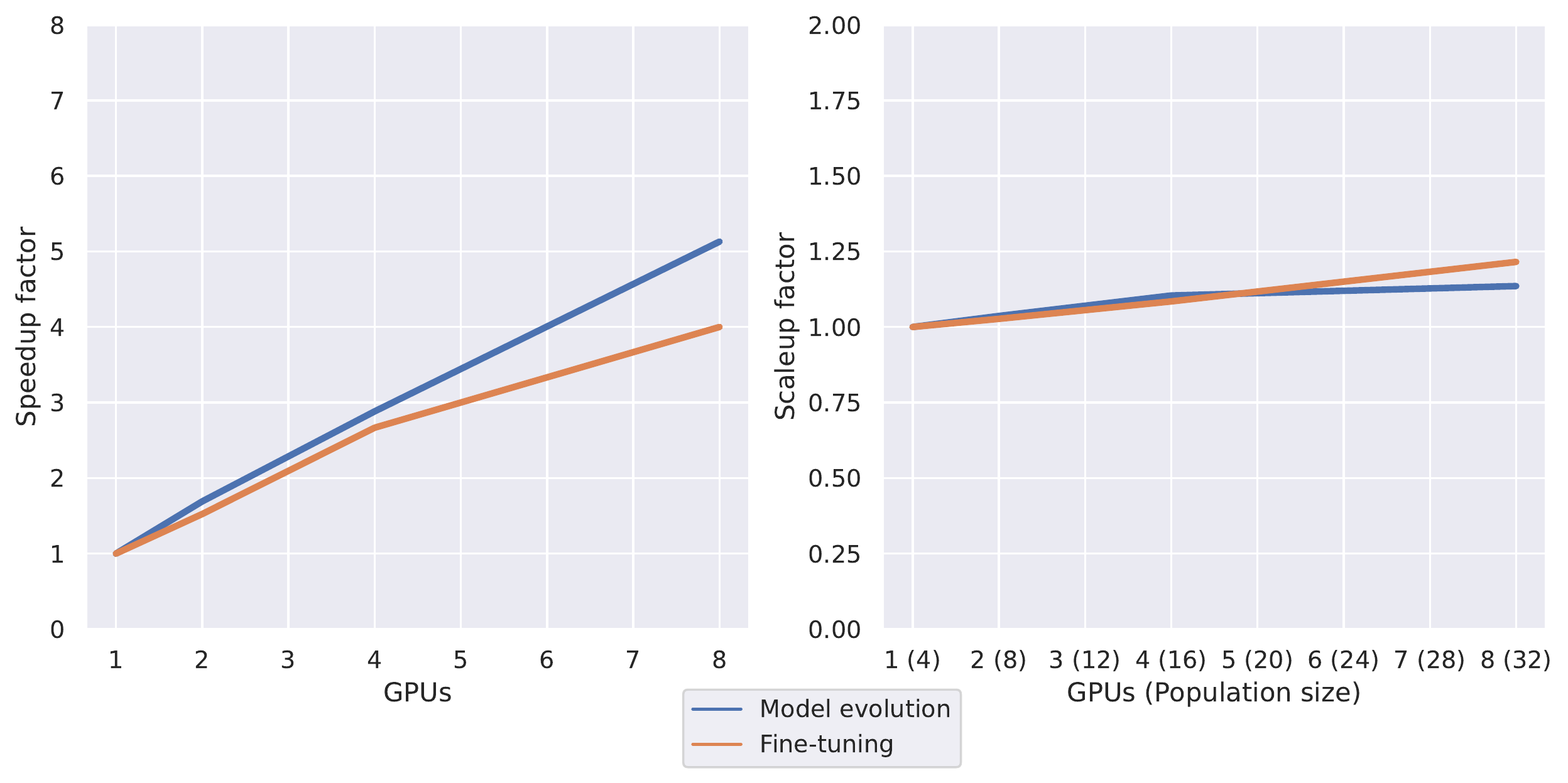}
\caption{Scalability results showing speedup (\textit{left}) and scaleup (\textit{right}) factors for both model evolution and fine-tuning levels with an increasing number of GPUs and population size (for scaleup only) on the \textit{x}-axis.}
\label{fig:speedup}
\end{figure}

It is also noteworthy that the execution time can be easily be reduced by adding more GPGPUs, since each model can be trained and evaluated independently on a separate GPGPU. Results in Figure \ref{fig:speedup} show that a speedup of up to 4x can be obtained for the fine-tuning level, and up to 5x for the model evolution level \footnote{The hyperparameter values specification for the scalability experiment are reported in Table \ref{tab:genetic_params_all} and \ref{tab:alg_params}}. The results also highlight that model evolution level of the AD-NEv framework exhibits a linear speedup with an increasing number of GPUs. The fine tuning process also benefits from the execution on multiple GPUs, achieving comparable performances with respect to the model evolution level. 
Figure \ref{fig:speedup} also shows results in terms of scaleup, where problem complexity (population size) and computational resources (GPUs) increase at the same time, i.e. from a population size of 4 with 1 GPU, to a population size of 32 with 8 GPUs. Results show a remarkable performance which positions very close to the ideal curve, i.e. a constant scaleup of 1 across all configurations.
Both speedup and scaleup results support our assumption that multiple GPGPUs can significantly reduce the execution time of the AD-NEv framework, effectively distributing the training and evaluation of multiple models on different GPUs, thus resulting in a significant speedup. Overall, the AD-NEv framework can be effectively scaled, supporting the efficient optimization of a large number of models (\textbf{RQ4}). 

    The models extracted after the neuroevolution process differ depending on the specific dataset and subspace. We present the complete architecture of the best model for the best subspace in every dataset in the supplementary material. The final models for the SWAT and WADI datasets consist of three convolutional layers in the encoder and the decoder parts of the model. However, the SWAT model presents a higher number of channels. The models for MSL and SMAP consist of six convolutional layers in the encoder and decoder.
    After the first two levels of the framework and before fine-tuning, the final training was run with 120 epochs in the case of the WADI dataset, 90 epochs for SWAT, and 250 for SMAP and MSL.
    
    



It is noteworthy that presented AD-NEv framework requires a smaller number of computational units than the state-of-the-art ensemble method in \cite{garg}, which achieves the best anomaly detection performance on multivariate time series benchmarks. While the model in \cite{garg} requires a number of submodels equal to the number of features, $S=M$,  AD-NEv produces an ensemble model with $K$ submodels working on $K$ subspaces, $K<<M$. The complexity and memory reduction of our solution is $\simeq\frac{K}{M}$ (the impact of the input size is negligible).

\section{Convergence study}
\label{sec:ablation_study}
This section presents the impact of neuroevolution parameters on model's accuracy. A detailed analysis for the most complex dataset (WADI) is presented in the following, focusing on a selection of relevant model parameters. Regarding the SMAP and the MSL datasets,  a separate model was generated and tuned for each entity (see Section \ref{sec:datasts}). For this reason, we do not present partial results for those datasets. It should be noted that, for all datasets, the behavior is consistent  with that observed with WADI.


\subsection{Convergence of the model evolution process}
\begin{figure}[H]
    \small
      \centering
      \includegraphics[scale=0.55]{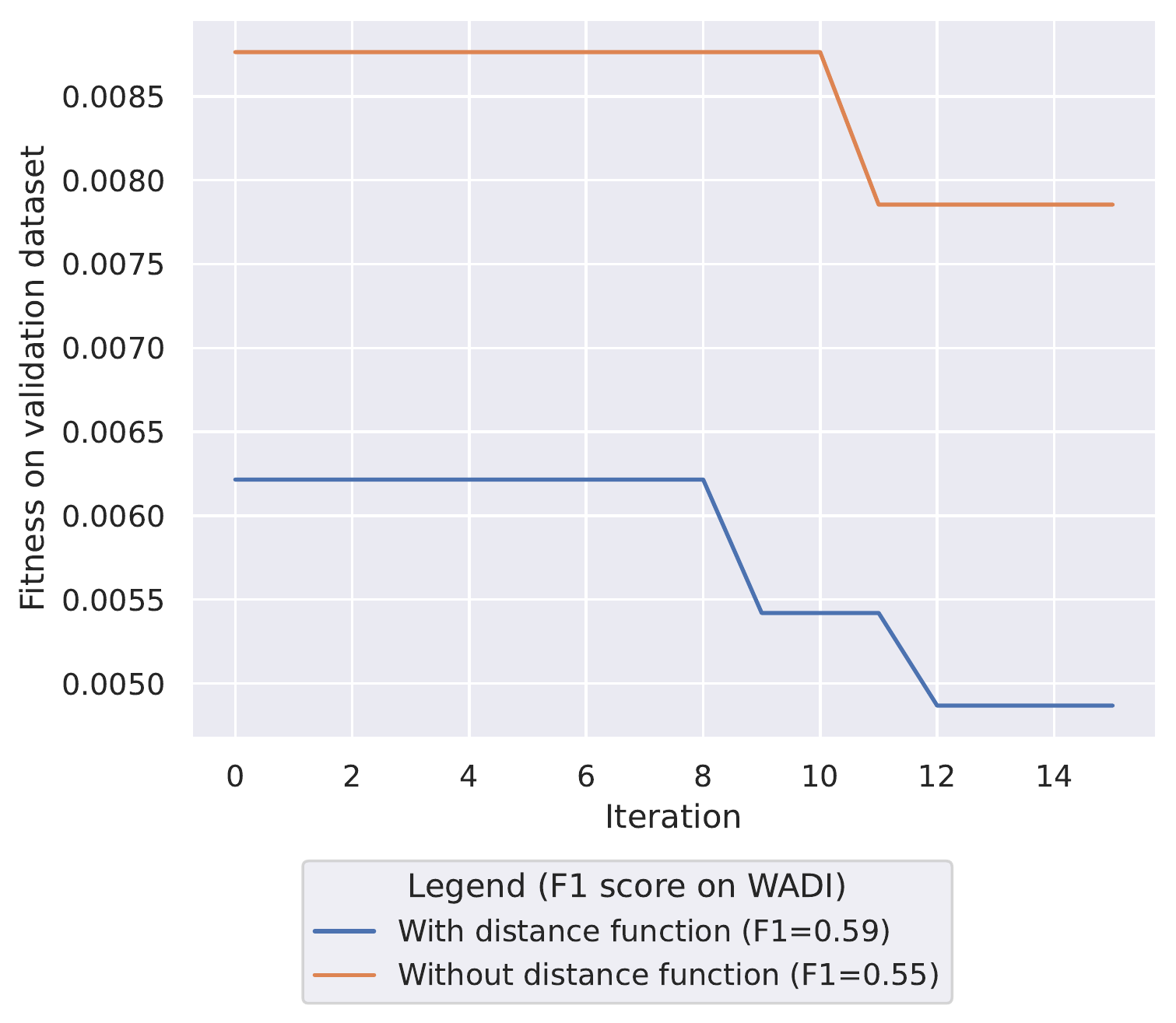}
        \caption{Convergence plot for the WADI dataset during the models evolution for population size 24. The diversity in selection process introduced by distance function improves not only convergence, but also F1 score on test dataset (F1 0.59 in comparison to 0.55 without distance function).}
        \label{fig:build_model}
    \end{figure}

Figure \ref{fig:build_model} presents the fitness on the validation dataset (not containing anomalies) of the best model in each iteration during model evolution process for version: \textit{i)} with the distance function for model evolution $d_F(F_i, F_j)$ introduced in Section \ref{subsec:models_evolution} - \textit{blue} and \textit{ii)} without the distance function - \textit{orange}. The distance function aims at achieving model diversity during the model evolution process. It could be observed that, in the scenario where the distance function is adopted, the value of the loss function was persisting on the same level over the first eight iterations. 
During the following eight iterations, the fitness value was improved twice. Finally, the model with the lowest value of loss function was evaluated on the test dataset achieving $F_1$ score equal to $0.59$. In the case of the scenario without diversity introduced by the distance function, the stagnation was persistent over 10 iterations, and it improved only once in the following iterations. Moreover, in this case, the loss value was always higher than the scenario with distance function. The best model evolved without distance function achieved the final $F_1$ score of $0.54$. We can clearly see that introducing the distance function $d_F(F_i, F_j)$ ensuring diversity to the population during evolution improved not only the convergence, but also the anomaly detection performance (\textbf{RQ5}).
Table \ref{tab:f1_depdends_on_pop_size} presents $F_1$ scores for WADI and SWAT on the test dataset, with different values of the population size parameter. As expected, the most effective models were extracted with the highest value for this parameter (24), showing the contribution of a large population of models during the evolution process\footnote{The largest population size was set to 24 due to time limitations given the available computation resources.}



\begin{table}[]
\centering
\caption{$F1$ score on WADI and SWAT datasets after evolution model process depending on population size.}
\begin{tabular}{|c|c|c|}
\hline
\textbf{Population size} & \textbf{$F_1$ on WADI} & \textbf{$F_1$ on SWAT} \\ \hline
\textbf{8}               & 0.54                  & 0.79                  \\ \hline
\textbf{16}              & 0.56                  & 0.80                  \\ \hline
\textbf{24}              & 0.59                  & 0.81                  \\ \hline
\end{tabular}
\label{tab:f1_depdends_on_pop_size}
\end{table}

\subsection{Convergence of the non-gradient fine-tuning process}

\label{sec:convergence_fine-tuning}

Results in Table \ref{tab:fine_tuning_convergence} show the average convergence and the $F_1$ obtained on the WADI dataset using different configurations for $N_p$, $p_m$, and $\tau$ parameters. 
Results show that with an increasing value of the population size, the algorithm required a reduced number of iterations to achieve a zero value of FP (8 iterations on average for $N_P=8$, and 4.92 on average iterations for $N_P=24$). Moreover, a higher number of $N_P$ also translates to an improved anomaly detection performance, resulting in an increased $F_1$ value.

\begin{table}[h]
\centering
\caption{Fine tuning convergence (number of iterations) and $F_1$ obtained with varying values of $N_P$, $p_m$, and $\tau$.}
\begin{tabular}{|c|c|c|c|c|}
\hline
\textbf{$N_P$} & \textbf{$p_m$} & \textbf{$\tau$}  & \textbf{$F_1$ on WADI} & \textbf{Avg convergence} \\
 & &  &  &  (iterations) \\\hline
\textbf{8}               & 0.02                  & $\frac{1}{256}$    & 0.60   & 8                 \\ \hline

\textbf{16}               & 0.02                  & $\frac{1}{256}$    & 0.61   &  5.82                 \\ \hline
\textbf{24}               & 0.02                  & $\frac{1}{256}$   & 0.62   & 4.92                 \\ \hline
\hline
\textbf{24}               & 0.02                  & $\frac{1}{128}$    & 0.59   &  2.67                 \\ \hline

\textbf{24}               & 0.02                  & $\frac{1}{512}$    & 0.62   & 12.8                 \\ \hline

\textbf{24}               & 0.05                  & $\frac{1}{256}$    & 0.53   &  21.33                 \\ \hline

\end{tabular}
\label{tab:fine_tuning_convergence}
\end{table}    

\begin{figure}[h]
    \small
      \centering
      \includegraphics[scale=0.6]{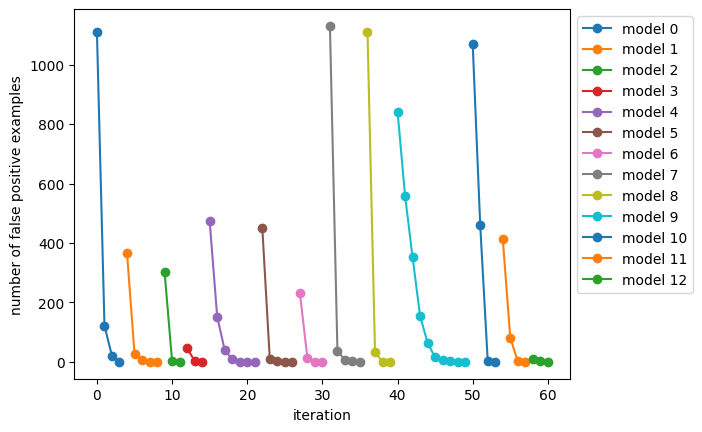}
        \caption{Fine-tuning convergence
        in terms of a different number false-positives (FP) for the best generated model in each iteration. The experiments were conducted on the WADI dataset with a population size of 24, $\tau=\frac{1}{256}$, and a mutation probability of 0.02. We also report F1 for each converged fine-tuned model.}
        \label{fig:convergence_pop_size_24_iter_64}
    \end{figure}

Figure 
\ref{fig:convergence_pop_size_24_iter_64} represents a detailed example for row 3 in Table \ref{tab:fine_tuning_convergence} (best configuration), showing the values of the fitness function. 
Additional plots showing the convergence trends of all models are reported in the Appendix.

Another perspective of results in Table \ref{tab:fine_tuning_convergence} is 
how fast the model was able to converge with different values of the mutation power $\tau$. Results (rows 3, 4 and 5) shows that the optimal model was obtained faster when this parameter had a higher value $(\frac{1}{128})$, but it does not have an impact on the final $F_1$ score. Moreover, using a large value for $\tau$ allows the process to converge faster.
It is worth noting that reducing the value of the $\tau$ parameter to $\tau=\frac{1}{512}$ allowed us to obtain the same $F_1$ score as $\tau=\frac{1}{128}$ but the algorithm required a higher number of iterations to converge. We argue that generating a compact set of models may be preferred over a large number of models to reduce the time impact of models evaluation on testing data.



Finally, one interesting aspect worth analysis is how increasing the value of mutation probability $p_m$ from 2\% ($p_m=0.02$) to 5\% ($p_m=0.05$) resulted in a decrease of efficiency for the algorithm. In this case, the best achieved $F_1$ score was 0.53, which is worse than the base $F_1$ value of 0.59. 
Comparing row 3 and 5, it could be observed that when too many weights are modified during a single iteration, the convergence process is slower. Overall, results show that the convergence process for the fine-tuning level is stable and efficient with respect to selected ranges of values of population size, $\tau$, and mutation probability parameters (\textbf{RQ5}).

{\color{red}


}

\section{Conclusions and future work}
\label{section:future_work}
In this paper, we proposed AD-NEv -- a novel multi-level framework for evolving ensemble deep learning auto-encoder architectures for multivariate anomaly detection. Its novelty consists in an efficient multi-level optimization that includes the evolution of input features subspaces, the model architecture for each subspace, and non-gradient fine-tuning. To further boost the anomaly detection performance, AD-NEv builds an ensemble model with voting to combine the outputs of single optimized models. 
We show that each introduced optimization component contributes to the achievement of the final model accuracy. An extensive experimental evaluation results show that the final ensemble model outperforms state-of-the-art anomaly detection models for multivariate time series data, while presenting a reasonable execution time, even with large datasets. AD-NEv's execution can be seamlessly scaled up by adding  computational resources. 
Directions for future work include the integration of the framework with redefined genetic operators for graph neural networks. Another possibility worth of investigation is the introduction of modified operators that enhance neural network architectures, such as different types of layers in a model or adding dense connections to each layer. 
Finally, we will further explore possible data distillation strategies to further reduce training data and reduce the execution time of the evolution level.

\bibliographystyle{unsrt}  
\bibliography{references}

\end{document}